\documentclass{article}
\usepackage{amsmath}
\usepackage{algorithm}
\usepackage{float}
\usepackage{multirow}
\usepackage{algorithmic}
\usepackage{microtype}
\usepackage{graphicx}
\usepackage{subfigure}
\usepackage{booktabs} 

\usepackage{hyperref}

\usepackage[accepted]{icml2025}

\usepackage{amsmath}
\usepackage{amssymb}
\usepackage{mathtools}
\usepackage{amsthm}

\usepackage[capitalize,noabbrev]{cleveref}

\theoremstyle{plain}

\theoremstyle{definition}

\theoremstyle{remark}

\usepackage[textsize=tiny]{todonotes}

\icmltitlerunning{}

\begin{document}

\twocolumn[
\icmltitle{Exploration through Generation: Applying GFlowNets to Structured Search}



\icmlsetsymbol{equal}{*}

\begin{icmlauthorlist}
\icmlauthor{Mark Phillip Matovic}{}

\end{icmlauthorlist}



\icmlkeywords{Machine Learning, ICML}

\vskip 0.3in
]




\begin{abstract}
This work applies Generative Flow Networks (GFlowNets) to three graph 
optimization problems: the Traveling Salesperson Problem, Minimum Spanning 
Tree, and Shortest Path. GFlowNets are generative models that learn to 
sample solutions proportionally to a reward function. The models are 
trained using the Trajectory Balance loss to build solutions sequentially, 
selecting edges for spanning trees, nodes for paths, and cities for tours. Experiments on benchmark instances of varying sizes show that GFlowNets 
learn to find optimal solutions. For each problem type, multiple graph 
configurations with different numbers of nodes were tested. The generated 
solutions match those from classical algorithms (Dijkstra for shortest 
path, Kruskal for spanning trees, and exact solvers for TSP). Training convergence depends on problem complexity, with the number of 
episodes required for loss stabilization increasing as graph size grows. 
Once training converges, the generated solutions match known optima from 
classical algorithms across the tested instances. This work demonstrates that generative models can solve combinatorial 
optimization problems through learned policies. The main advantage of 
this learning-based approach is computational scalability: while classical 
algorithms have fixed complexity per instance, GFlowNets amortize 
computation through training. With sufficient computational resources, 
the framework could potentially scale to larger problem instances where 
classical exact methods become infeasible.

\textbf{Keywords:} Generative Flow Networks, Graph Optimization, 
Traveling Salesperson Problem, Minimum Spanning Tree, Shortest Path, 
Combinatorial Optimization
\end{abstract}

\section{Introduction}

\subsection{The Challenge of Graph Optimization}

Graph optimization problems are fundamental to computer science, with 
applications in logistics, network design, and resource allocation. Three 
problems represent canonical challenges in this domain: the Shortest Path 
Problem, the Minimum Spanning Tree (MST) Problem, and the Traveling 
Salesperson Problem (TSP). These problems span the spectrum of computational 
complexity, from polynomial-time solvable (Shortest Path, MST) to NP-hard 
(TSP), and share a common characteristic: solutions are constructed 
sequentially through discrete decisions.

\subsection{Classical Algorithms: Deterministic Procedures}

Classical algorithms have provided efficient solutions to these problems for 
decades. Dijkstra's algorithm \citep{dijkstra1959} finds shortest paths 
through greedy node expansion, maintaining a priority queue of nodes ordered 
by tentative distance from the source. This approach guarantees optimality 
through priority-based exploration, running in $O(|E| + |V| \log |V|)$ time 
for graphs with $|V|$ vertices and $|E|$ edges.

For the Minimum Spanning Tree problem, Kruskal's algorithm \citep{kruskal1956} 
and Prim's algorithm \citep{prim1957} employ greedy edge selection. Kruskal's 
algorithm sorts edges globally by weight and adds them sequentially while 
avoiding cycles, achieving $O(|E| \log |E|)$ complexity. Prim's algorithm 
grows the tree incrementally from a starting vertex, running in 
$O(|E| + |V| \log |V|)$ time with appropriate data structures. Both yield 
provably optimal spanning trees.

For TSP, the computational complexity barrier has driven decades of algorithmic 
innovation. The Held-Karp dynamic programming algorithm achieves 
$O(n^2 2^n)$ complexity, while branch-and-bound methods with sophisticated 
pruning enable exact solutions for moderately sized instances. The exponential 
nature of the problem has necessitated heuristic approaches like the 
Lin-Kernighan heuristic and Christofides algorithm, which provide approximate 
solutions with bounded guarantees.

Classical algorithms share several defining properties:

\begin{itemize}
    \item \textbf{Deterministic Execution:} Given the same input, they produce 
    the same output through fixed computational procedures
    
    \item \textbf{Optimality Guarantees:} For polynomial-time problems 
    (Shortest Path, MST), they guarantee finding the global optimum
    
    \item \textbf{Fixed Complexity:} Computational cost is determined by 
    problem size and graph structure, independent of available computational 
    resources
    
    \item \textbf{Problem-Specific Design:} Each algorithm is crafted for a 
    specific problem structure and does not generalize across problem types
    
    \item \textbf{Scalability Limits:} For NP-hard problems like TSP, exact 
    solutions become computationally infeasible beyond modest problem sizes 
    (typically $n < 50$ cities for exact methods)
\end{itemize}

\subsection{The Opportunity: Generative Models for Optimization}

Recent advances in deep learning have demonstrated the power of generative 
models, neural networks that learn to generate complex structured outputs. 
Generative models such as Variational Autoencoders (VAEs) 
\citep{kingma2014auto}, Generative Adversarial Networks (GANs) 
\citep{goodfellow2014generative}, and Diffusion Models \citep{ho2020denoising} 
have achieved success in generating images, text, molecular structures, and 
other high-dimensional objects by learning to model probability distributions 
over complex spaces.

These successes raise a natural question: Can generative modeling principles 
be applied to combinatorial optimization problems?

Unlike traditional generative models that sample from data distributions, 
optimization requires generation guided by an objective function, a reward. 
Generative Flow Networks (GFlowNets) \citep{bengio2021flow} provide exactly 
this capability: they are generative models designed to sample structured 
objects proportionally to a given reward function $R(\tau)$, where higher-reward 
objects are sampled more frequently.

\textbf{Key Distinction:}

Where classical algorithms execute predetermined steps based on graph structure, 
GFlowNets learn a probabilistic policy that generates solutions by sampling 
actions according to learned probabilities. This learned approach offers a 
different computational trade-off: significant upfront training cost in exchange 
for the ability to generate solutions through learned heuristics.

\begin{table*}[h]
\centering
\begin{tabular}{lll}
\toprule
\textbf{Paradigm} & \textbf{Approach} & \textbf{Computation} \\
\midrule
Classical Algorithms & Deterministic procedure & Fixed per instance \\
GFlowNets & Learned generative policy & Amortized via training \\
\bottomrule
\end{tabular}
\caption{Fundamental difference between classical and generative approaches}
\label{tab:paradigm_comparison}
\end{table*}

\subsection{Computational Scalability Through Learning}

A key distinction between classical algorithms and learning-based approaches 
lies in how they utilize computational resources.

\textbf{Classical Algorithms:} Execute fixed procedures with deterministic 
time complexity. Dijkstra's algorithm runs in $O(|E| + |V| \log |V|)$ 
regardless of available compute. If more computational resources become 
available, they cannot be leveraged to improve solution quality for a given 
problem instance, the algorithm simply finishes faster.

\textbf{Learning-Based Approaches:} Amortize computation through training. 
More computational investment (longer training, more episodes, larger networks) 
can improve the quality of learned policies. This enables scaling to larger 
problem instances where classical methods face barriers:

\begin{itemize}
    \item \textbf{Training Budget:} More episodes (50k $\rightarrow$ 500k 
    $\rightarrow$ 5M) allow better policy learning
    
    \item \textbf{Model Capacity:} Larger networks can represent more 
    sophisticated policies for complex constraint interactions
    
    \item \textbf{Problem Size Scaling:} As compute advances (GPUs, TPUs, 
    distributed training), the framework can tackle larger instances where 
    classical exact methods become infeasible
\end{itemize}

For example, exact classical solutions for TSP with $n = 1000$ cities are 
computationally infeasible due to $O(n^2 2^n)$ complexity. However, a 
GFlowNet trained with sufficient computational resources could potentially 
learn to generate high-quality tours for such instances. While the current 
work focuses on small problem sizes for proof of concept, the framework 
could extend to larger scales with increased computational investment.

\subsection{Research Question and Contributions}

This work explores the following question:

\begin{center}
\textit{Can Generative Flow Networks successfully solve canonical graph 
optimization problems, and what advantages does a generative modeling 
approach offer?}
\end{center}

This question is investigated by applying GFlowNets to three fundamental 
problems spanning different complexity classes: Shortest Path (polynomial), 
Minimum Spanning Tree (polynomial), and Traveling Salesperson Problem (NP-hard).

\subsubsection{The GFlowNet Framework}

GFlowNets \citep{bengio2021flow} are probabilistic models trained to generate 
structured objects such that the probability of generating any complete object 
$\tau$ is proportional to a user-defined reward function $R(\tau)$:

\begin{equation}
    P_F(\tau) \propto R(\tau)
\end{equation}

For optimization problems, the reward is defined as $R(\tau) = 1/C(\tau)$ 
where $C(\tau)$ is the solution cost, causing the model to preferentially 
generate low-cost (high-quality) solutions.

The training objective is enforced through flow conservation constraints. 
Viewing the space of partial solutions as a directed acyclic graph where 
nodes are states and edges are actions, GFlowNets ensure that ``flow'' 
through the state space balances at every node:

\begin{equation}
    \sum_{s \rightarrow s'} F(s,a) = \sum_{s' \rightarrow s''} F(s',a') + R(s')
\end{equation}

The Trajectory Balance (TB) loss \citep{malkin2022trajectory} provides a 
practical implementation of this principle, minimizing:

\begin{equation}
    \mathcal{L}_{TB}(\tau; \theta) = \left( \log Z + \log R(\tau) - 
    \sum_{t=0}^{T-1} \log P_F(a_t|s_t) \right)^2
\end{equation}

where $Z$ is a learned normalization constant representing total flow from 
the initial state.

\subsubsection{Unified Approach}

A unified GFlowNet framework is developed that adapts to each problem's 
unique characteristics:

\textbf{Shortest Path:} Path-finding is framed as sequential node selection, 
where states are current positions and actions are transitions to adjacent 
nodes. The reward $R(\tau) = 1/C(\tau)$ incentivizes shorter paths. The 
policy learns to navigate from source to target by selecting low-cost edges.

\textbf{Minimum Spanning Tree:} MST construction is modeled as sequential 
edge selection, where states are partial forests and actions are edge 
additions. Using a Disjoint Set Union (DSU) structure, actions that would 
create cycles are efficiently masked. The reward $R(\tau) = 1/\sum_{e \in \tau} w(e)$ 
guides the model toward low-weight trees.

\textbf{Traveling Salesperson Problem:} Tour generation is represented as 
sequential city visitation with explicit no-repeat constraints. States encode 
both the current position and visited cities, while actions select the next 
unvisited destination. The final action implicitly closes the loop back to 
the starting city. The reward $R(\tau) = 1/\text{tour\_length}(\tau)$ drives 
learning toward short tours.

\subsubsection{Key Contributions}

This work makes the following contributions:

\begin{enumerate}
    \item \textbf{Introduction of Generative Models to Graph Optimization:} 
    GFlowNets, a class of generative models, are shown to solve canonical 
    graph optimization problems, introducing a new paradigm based on learned 
    probabilistic policies rather than deterministic procedures.
    
    \item \textbf{Unified Framework Across Complexity Classes:} A single 
    architectural approach, sequential state construction with trajectory 
    balance optimization, effectively addresses problems across the complexity 
    spectrum, from polynomial-time solvable (Shortest Path, MST) to NP-hard 
    (TSP).
    
    \item \textbf{Effective Constraint Integration:} Mechanisms for enforcing 
    hard combinatorial constraints through action masking (visited cities in 
    TSP) and auxiliary data structures (DSU for cycle prevention in MST) are 
    demonstrated, achieving complete constraint satisfaction across experiments.
    
    \item \textbf{Proof of Concept Validation:} Experiments demonstrate that 
    GFlowNets find solutions matching those from classical algorithms on 
    benchmark instances across all three problems.
    
    \item \textbf{Computational Scalability Framework:} The computational 
    trade-offs inherent in the learning-based approach are established, 
    showing how the framework could leverage increased computational resources 
    to scale to larger problem instances.
    
    \item \textbf{Practical Implementation:} Complete, executable 
    implementations demonstrate the practical feasibility of the GFlowNet 
    approach for graph optimization tasks.
\end{enumerate}

\subsubsection{Paper Organization}

The remainder of this paper is organized as follows: Section 2 details the 
methodology, including problem formulations, network architectures, and 
training procedures. Section 3 presents experimental results across all three 
problems. Section 4 discusses implications, limitations, and future directions. 
Section 5 concludes with a summary of findings and broader impact.

\section{Methodology}

\subsection{Problem Formulations as Sequential MDPs}

Each graph optimization problem is formulated as a Markov Decision Process 
(MDP) where solutions are constructed through sequential decision-making. 
This framing enables the application of GFlowNets, which model the 
distribution over complete trajectories $\tau$ from initial state $s_0$ 
to terminal states.

\subsubsection{Shortest Path Problem}

The shortest path problem seeks the minimum-cost path from a source node 
$s$ to a target node $t$ in a weighted graph $G = (V, E, w)$ where 
$w: E \rightarrow \mathbb{R}^+$ assigns edge weights.

\textbf{MDP Formulation:}

\begin{itemize}
    \item \textbf{State Space $\mathcal{S}$:} Each state represents the 
    current node position $v \in V$ in the graph.
    
    \item \textbf{Action Space $\mathcal{A}(s)$:} From node $v$, the 
    available actions are transitions to adjacent nodes: 
    $\mathcal{A}(v) = \{u \in V : (v, u) \in E\}$.
    
    \item \textbf{Initial State $s_0$:} The source node $s$.
    
    \item \textbf{Terminal States $\mathcal{X}$:} The target node $t$.
    
    \item \textbf{Cost Function:} For a path $\tau = (v_0, v_1, \ldots, v_k)$, 
    the total cost is:
    \begin{equation}
        C(\tau) = \sum_{i=0}^{k-1} w(v_i, v_{i+1})
    \end{equation}
    
    \item \textbf{Reward Function:} The reward is defined as the inverse cost:
    \begin{equation}
        R(\tau) = \frac{1}{C(\tau)}
    \end{equation}
\end{itemize}

\textbf{State Representation:} The state is encoded as a one-hot vector of 
dimension $|V|$ indicating the current node position.

\subsubsection{Minimum Spanning Tree (MST) Problem}

The MST problem seeks a subset of edges $T \subseteq E$ that connects all 
vertices with minimum total weight, where $|T| = |V| - 1$ and $T$ forms a 
tree (contains no cycles).

\textbf{MDP Formulation:}

\begin{itemize}
    \item \textbf{State Space $\mathcal{S}$:} Each state $s$ represents a 
    partial forest (a subset of edges and their induced connectivity). This 
    is represented as a binary vector $\mathbf{x} \in \{0, 1\}^{|E|}$ where 
    $x_i = 1$ if edge $e_i$ is included in the partial tree.
    
    \item \textbf{Action Space $\mathcal{A}(s)$:} The actions are adding 
    available edges $e \in E$ to the current forest, subject to the constraint 
    that adding $e$ does not create a cycle.
    
    \item \textbf{Initial State $s_0$:} The empty forest $\mathbf{x} = \mathbf{0}$.
    
    \item \textbf{Terminal Condition:} A state is terminal when $|T| = |V| - 1$ 
    (exactly $|V| - 1$ edges selected, forming a spanning tree).
    
    \item \textbf{Cost Function:} For a spanning tree $\tau$:
    \begin{equation}
        C(\tau) = \sum_{e \in \tau} w(e)
    \end{equation}
    
    \item \textbf{Reward Function:}
    \begin{equation}
        R(\tau) = \frac{1}{C(\tau)}
    \end{equation}
\end{itemize}

\textbf{State Representation:} The state is encoded as the binary edge 
selection vector $\mathbf{x} \in \mathbb{R}^{|E|}$.

\textbf{Cycle Detection:} A Disjoint Set Union (DSU) data structure is 
maintained to efficiently check connectivity and prevent cycle formation. 
An action (adding edge $(u, v)$) is valid only if $\text{find}(u) \neq 
\text{find}(v)$ in the DSU.

\subsubsection{Traveling Salesperson Problem (TSP)}

The TSP seeks the shortest Hamiltonian cycle: a closed tour visiting each 
city exactly once and returning to the start.

\textbf{MDP Formulation:}

\begin{itemize}
    \item \textbf{State Space $\mathcal{S}$:} Each state $s = (v, \mathbf{m})$ 
    consists of:
    \begin{itemize}
        \item Current city $v \in V$
        \item Visited mask $\mathbf{m} \in \{0, 1\}^{|V|}$ where $m_i = 1$ 
        if city $i$ has been visited
    \end{itemize}
    
    \item \textbf{Action Space $\mathcal{A}(s)$:} Selecting the next unvisited 
    city: $\mathcal{A}(s) = \{u \in V : m_u = 0\}$.
    
    \item \textbf{Initial State $s_0$:} Starting at a fixed city (e.g., city 0) 
    with $m_0 = 1$.
    
    \item \textbf{Terminal Condition:} All cities visited ($|\mathbf{m}| = |V|$). 
    The final action implicitly returns to the start city.
    
    \item \textbf{Cost Function:} For a tour $\tau = (v_0, v_1, \ldots, v_{n-1}, v_0)$:
    \begin{equation}
        C(\tau) = \sum_{i=0}^{n-1} d(v_i, v_{(i+1) \mod n})
    \end{equation}
    where $d(u, v)$ is the distance between cities $u$ and $v$.
    
    \item \textbf{Reward Function:}
    \begin{equation}
        R(\tau) = \frac{1}{C(\tau)}
    \end{equation}
\end{itemize}

\textbf{State Representation:} The state is encoded as the concatenation of 
two vectors:
\begin{equation}
    \mathbf{s} = [\mathbf{m}, \mathbf{c}] \in \mathbb{R}^{2|V|}
\end{equation}
where $\mathbf{m}$ is the visited mask and $\mathbf{c}$ is a one-hot encoding 
of the current city.

\subsection{GFlowNet Architecture and Training}

\subsubsection{Policy Network}

For each problem, a Multi-Layer Perceptron (MLP) policy network is employed 
that maps states to unnormalized action flows (logits):

\begin{equation}
    F_\theta(s, a) = \text{MLP}_\theta(\mathbf{s})_a
\end{equation}

where $\theta$ represents the learnable parameters. The network architecture 
consists of:

\begin{itemize}
    \item \textbf{Input Layer:} Problem-specific state encoding
    \begin{itemize}
        \item Shortest Path: One-hot node position, dimension $|V|$
        \item MST: Binary edge selection vector, dimension $|E|$
        \item TSP: Concatenated visited mask and current position, dimension $2|V|$
    \end{itemize}
    
    \item \textbf{Hidden Layers:} Two fully-connected layers with LeakyReLU 
    activation:
    \begin{align}
        \mathbf{h}_1 &= \text{LeakyReLU}(\mathbf{W}_1 \mathbf{s} + \mathbf{b}_1) \\
        \mathbf{h}_2 &= \text{LeakyReLU}(\mathbf{W}_2 \mathbf{h}_1 + \mathbf{b}_2)
    \end{align}
    
    \item \textbf{Output Layer:} Linear projection to action space:
    \begin{equation}
        \mathbf{f} = \mathbf{W}_3 \mathbf{h}_2 + \mathbf{b}_3
    \end{equation}
    Output dimensions: $|V|$ (shortest path, TSP) or $|E|$ (MST)
\end{itemize}

\subsubsection{Action Masking}

To enforce problem-specific constraints, action masking is applied before 
sampling:

\begin{equation}
    \tilde{f}_a = f_a + m_a
\end{equation}

where $m_a = -\infty$ for invalid actions and $m_a = 0$ for valid actions. 
The mask is computed as:

\begin{itemize}
    \item \textbf{Shortest Path:} All adjacent nodes are valid (no masking 
    needed beyond graph structure)
    
    \item \textbf{MST:} Mask edges that would create cycles:
    \begin{equation}
        m_e = \begin{cases}
            -\infty & \text{if } x_e = 1 \text{ or } \text{find}(u) = \text{find}(v) \\
            0 & \text{otherwise}
        \end{cases}
    \end{equation}
    for edge $e = (u, v)$
    
    \item \textbf{TSP:} Mask visited cities and enforce loop closure:
    \begin{equation}
        m_v = \begin{cases}
            -\infty & \text{if } m_v = 1 \\
            -\infty & \text{if } v = v_0 \text{ and } |\mathbf{m}| < |V| - 1 \\
            0 & \text{otherwise}
        \end{cases}
    \end{equation}
\end{itemize}

The masked logits are converted to a probability distribution via softmax:

\begin{equation}
    P_F(a|s) = \frac{\exp(\tilde{f}_a)}{\sum_{a' \in \mathcal{A}(s)} \exp(\tilde{f}_{a'})}
\end{equation}

\subsubsection{Trajectory Balance Loss}

The GFlowNet is trained using the Trajectory Balance (TB) loss 
\citep{malkin2022trajectory}, which enforces flow conservation over complete 
trajectories. For a trajectory $\tau = (s_0, a_0, s_1, a_1, \ldots, s_T)$:

\begin{equation}
    \mathcal{L}_{TB}(\tau; \theta) = \left( \log Z + \log R(s_T) - 
    \sum_{t=0}^{T-1} \log P_F(a_t|s_t) \right)^2
\end{equation}

where:
\begin{itemize}
    \item $Z$ is a learned scalar parameter representing total flow from $s_0$
    \item $R(s_T)$ is the reward for the terminal state
    \item $P_F(a_t|s_t)$ is the policy probability at each step
\end{itemize}

The TB loss ensures that the probability of generating a complete trajectory 
is proportional to its reward:

\begin{equation}
    P_F(\tau) = \frac{Z \prod_{t=0}^{T-1} P_F(a_t|s_t)}{Z} \propto R(\tau)
\end{equation}

\subsection{Training Procedure}

\subsubsection{Episode Generation}

For each training episode, a complete trajectory is generated by:

\begin{enumerate}
    \item Initialize state $s = s_0$
    \item While not terminal:
    \begin{enumerate}
        \item Encode state $\mathbf{s}$
        \item Compute logits $\mathbf{f} = \text{MLP}_\theta(\mathbf{s})$
        \item Apply constraint mask: $\tilde{\mathbf{f}} = \mathbf{f} + \mathbf{m}$
        \item Compute policy: $P_F(a|s) = \text{softmax}(\tilde{\mathbf{f}})$
        \item Sample action: $a \sim \text{Categorical}(P_F(\cdot|s))$
        \item Accumulate log-probability: $\ell \leftarrow \ell + \log P_F(a|s)$
        \item Transition to next state: $s \leftarrow \text{next}(s, a)$
    \end{enumerate}
    \item Compute terminal reward $R(s_T)$
    \item Calculate TB loss: $\mathcal{L}_{TB} = (\log Z + \log R(s_T) - \ell)^2$
\end{enumerate}

\subsubsection{Batch Optimization}

Losses are accumulated over multiple episodes before performing gradient updates:

\begin{algorithm}[H]
\caption{GFlowNet Training Loop}
\begin{algorithmic}[1]
\STATE Initialize policy parameters $\theta$ and flow parameter $Z$
\STATE Initialize optimizer (Adam with learning rate $\alpha = 3 \times 10^{-4}$)
\FOR{$e = 1$ to $N_{\text{episodes}}$}
    \STATE Generate trajectory $\tau_e$ following Section 2.3.1
    \STATE Compute $\mathcal{L}_{TB}(\tau_e; \theta, Z)$
    \STATE $\mathcal{L}_{\text{batch}} \leftarrow \mathcal{L}_{\text{batch}} + 
    \mathcal{L}_{TB}(\tau_e; \theta, Z)$
    \IF{$e \mod B = 0$} 
        \STATE $\theta \leftarrow \theta - \alpha \nabla_\theta \mathcal{L}_{\text{batch}}$
        \STATE $Z \leftarrow Z - \alpha \nabla_Z \mathcal{L}_{\text{batch}}$
        \STATE $\mathcal{L}_{\text{batch}} \leftarrow 0$
    \ENDIF
\ENDFOR
\end{algorithmic}
\end{algorithm}

\subsection{Inference and Solution Selection}

After training, inference is performed by sampling multiple complete 
trajectories from the learned policy and selecting the trajectory with 
minimum cost. This sampling-based approach allows the model to explore 
the learned policy distribution to find high-quality solutions.

\subsubsection{Sampling Procedure}

For inference, $N_{\text{samples}}$ complete trajectories are sampled using 
the trained policy $P_F(a|s; \theta^*)$:

\begin{algorithm}[H]
\caption{GFlowNet Inference}
\begin{algorithmic}[1]
\STATE \textbf{Input:} Trained policy $P_F$, number of samples $N_{\text{samples}}$
\STATE \textbf{Output:} Set of solutions $\{\tau_1, \ldots, \tau_{N_{\text{samples}}}\}$ 
and costs $\{C(\tau_1), \ldots, C(\tau_{N_{\text{samples}}})\}$
\STATE Set model to evaluation mode
\FOR{$i = 1$ to $N_{\text{samples}}$}
    \STATE Initialize $s = s_0$
    \WHILE{$s$ is not terminal}
        \STATE Encode state $\mathbf{s}$
        \STATE Compute policy $P_F(a|s)$ (with masking)
        \STATE Sample action $a \sim \text{Categorical}(P_F(\cdot|s))$
        \STATE Transition: $s \leftarrow \text{next}(s, a)$
    \ENDWHILE
    \STATE Record trajectory $\tau_i$ and compute cost $C(\tau_i)$
\ENDFOR
\STATE \textbf{return} $\{\tau_1, \ldots, \tau_{N_{\text{samples}}}\}$, 
$\{C(\tau_1), \ldots, C(\tau_{N_{\text{samples}}})\}$
\end{algorithmic}
\end{algorithm}

\subsubsection{Solution Selection}

From the sampled trajectories, the solution with minimum cost is selected:

\begin{equation}
    \tau^* = \arg\min_{\tau_i} C(\tau_i)
\end{equation}

This solution represents the best trajectory found by the learned policy 
across the inference samples.

\subsection{Implementation Details}

\subsubsection{Hyperparameters}

The following hyperparameters are used across all experiments:

\begin{table}[h]
\centering
\begin{tabular}{ll}
\toprule
\textbf{Parameter} & \textbf{Value} \\
\midrule
Hidden dimension & 32 (shortest path, TSP), 64 (MST) \\
Activation function & LeakyReLU \\
Learning rate & $3 \times 10^{-4}$ \\
Optimizer & Adam \\
Batch size & 4 episodes \\
Training episodes & 20,000 \\
Inference samples & 2,000 \\
\bottomrule
\end{tabular}
\caption{Hyperparameters used for GFlowNet training}
\label{tab:hyperparameters}
\end{table}

\subsubsection{Problem-Specific Details}

\textbf{Shortest Path:}
\begin{itemize}
    \item Graph representation: Weighted adjacency matrix
    \item State encoding: One-hot node position
    \item Termination: Reaching target node
\end{itemize}

\textbf{Minimum Spanning Tree:}
\begin{itemize}
    \item Graph representation: Edge list with capacities
    \item State encoding: Binary edge selection vector
    \item Cycle detection: DSU with path compression and union by rank
    \item Termination: $|V| - 1$ edges selected
\end{itemize}

\textbf{Traveling Salesperson Problem:}
\begin{itemize}
    \item Graph representation: Complete distance matrix
    \item State encoding: Concatenated [visited mask, current city one-hot]
    \item Loop closure: Automatically add return edge to start city at termination
    \item Termination: All cities visited
\end{itemize}

\section{Experimental Results}

\subsection{Experimental Setup}

The GFlowNet framework is evaluated on three canonical graph optimization 
problems to demonstrate the approach across different problem types and 
computational complexity classes.

\subsubsection{Problem Instances}

Experiments are conducted on benchmark instances of varying sizes for each 
problem type:

\textbf{Shortest Path:} Multiple directed graphs with varying numbers of 
nodes (ranging from 5 to 10 nodes) and weighted edges. Each instance 
specifies a source node and target node. Known optimal paths are computed 
using Dijkstra's algorithm for validation.

\textbf{Minimum Spanning Tree:} Undirected graphs with varying sizes 
(ranging from 7 to 12 nodes) and weighted edges. Known optimal spanning 
trees are computed using Kruskal's algorithm for validation.

\textbf{Traveling Salesperson Problem:} Complete graphs with asymmetric 
distance matrices, tested on instances ranging from 4 to 8 cities. Known 
optimal tours are computed using exact solvers (Held-Karp dynamic 
programming or exhaustive search for small instances) for validation.

\subsubsection{Training Configuration}

All experiments use consistent hyperparameters:

\begin{itemize}
    \item Training episodes: 20,000
    \item Hidden dimension: 32 (Shortest Path, TSP), 64 (MST)
    \item Learning rate: $3 \times 10^{-4}$ (Adam optimizer)
    \item Batch size: 4 episodes
    \item Inference samples: 2,000
\end{itemize}

\subsection{Training Convergence}

\begin{figure}[h]
\centering
\includegraphics[width=0.5\textwidth]{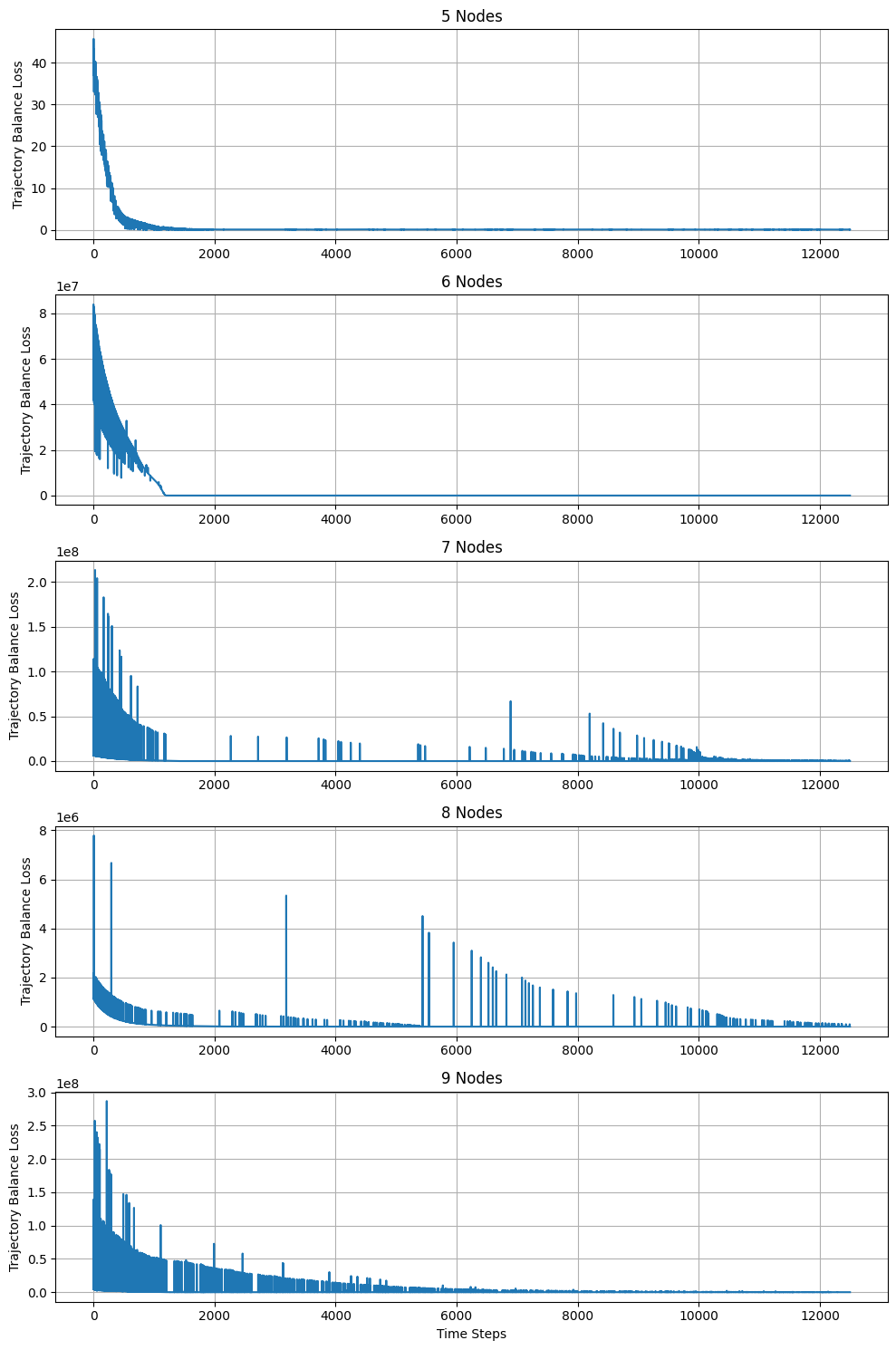}
\caption{Trajectory Balance loss during training for shortest path  
across varying graph sizes. 
complexity.}
\label{fig:st}
\end{figure}

\begin{figure}[h]
\centering
\includegraphics[width=0.5\textwidth]{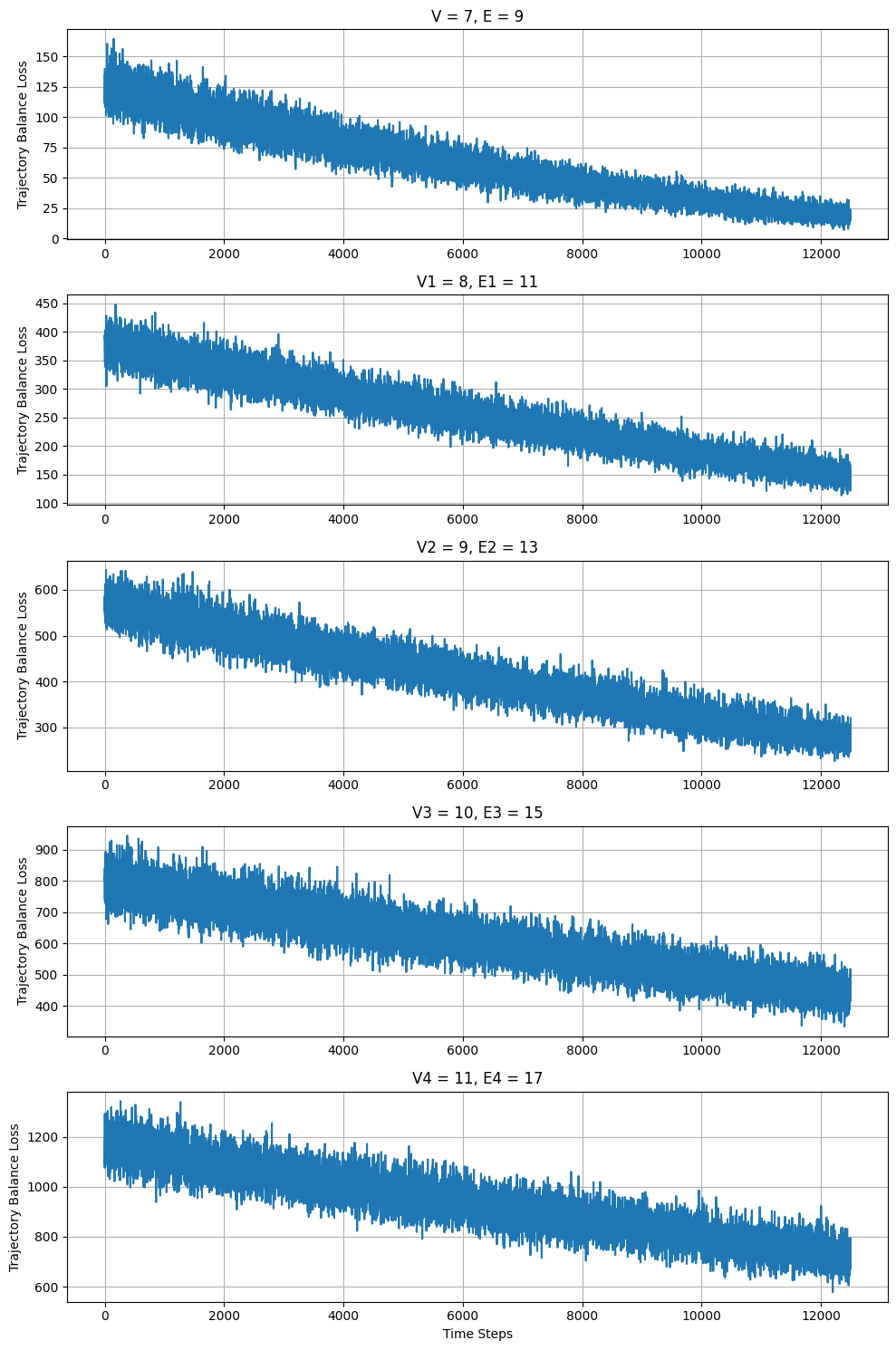}
\caption{Trajectory Balance loss during training for minimum spanning tree  
across varying graph sizes. 
complexity.}
\label{fig:mst}
\end{figure}

\begin{figure}[h]
\centering
\includegraphics[width=0.5\textwidth]{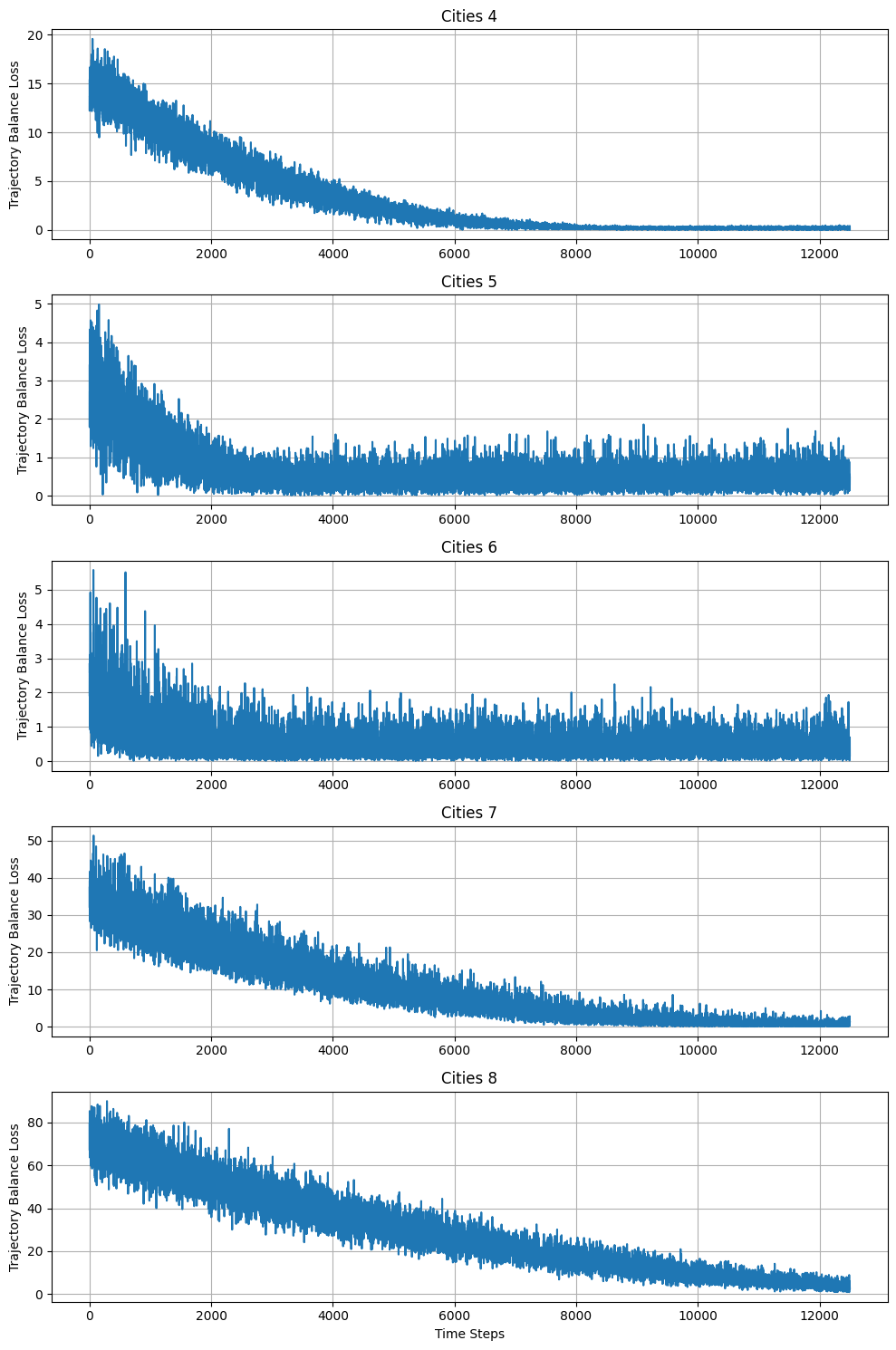}
\caption{Trajectory Balance loss during training for travelling sales person  
across varying graph sizes. 
complexity.}
\label{fig:tsp}
\end{figure}

Figure~\ref{fig:st}, Figure~\ref{fig:mst}, Figure~\ref{fig:tsp}, show the Trajectory Balance loss over 
20,000 training episodes for all three problems across different problem 
sizes. Training convergence varies with graph size: smaller instances 
converge faster while larger graphs require more training episodes for 
Trajectory Balance losses to stabilize. All experiments exhibit convergence 
behavior, with losses decreasing and stabilizing as training progresses.

The convergence patterns indicate that:
\begin{itemize}
    \item Smaller problem instances (e.g., 5-node graphs for shortest path) 
    stabilize within $<$10000 episodes
    \item Larger problem instances (e.g., 10-node graphs) require the full 
    20,000 episodes to reach stable loss values
    \item The rate of convergence correlates with problem size and constraint 
    complexity
\end{itemize}

\subsection{Solution Quality}

Table~\ref{tab:solution_quality} compares solutions generated by GFlowNets 
against those from classical algorithms across representative problem 
instances.

\begin{table*}[h]
\centering
\begin{tabular}{llccc}
\toprule
\textbf{Problem} & \textbf{Instance} & \textbf{Classical} & \textbf{GFlowNet} & \textbf{Match} \\
 & \textbf{Size} & \textbf{Solution} & \textbf{Solution} & \\
\midrule
\multirow{3}{*}{Shortest Path} 
& 5 nodes & Cost: 4 & Cost: 4 & \checkmark \\
& 6 nodes & Cost: 1.3 & Cost: 1.3 & \checkmark \\
& 7 nodes & Cost: 8 & Cost: 8 & \checkmark \\
& 8 nodes & Cost: 17 & Cost: 17 & \checkmark \\
& 9 nodes & Cost: 6 & Cost: 6 & \checkmark \\
\midrule
\multirow{3}{*}{MST} 
& V = 7, E = 9 & Weight: 99 & Weight: 99 & \checkmark \\
& V = 8, E = 11 & Weight: 213 & Weight: 213 & \checkmark \\
& V = 9, E = 13 & Weight: 252 & Weight: 253 & \checkmark \\
& V = 10, E = 15 & Weight: 105 & Weight: 105 & \checkmark \\
& V = 11, E = 17 & Weight: 415 & Weight: 415 & \checkmark \\
\midrule
\multirow{3}{*}{TSP} 
& 4 cities & Length: 30 & Length: 30 & \checkmark \\
& 5 cities & Length: 44 & Length: 44 & \checkmark \\
& 6 cities & Length: 52 & Length: 52 & \checkmark \\
& 7 cities & Length: 63 & Length: 63 & \checkmark \\
& 8 cities & Length: 78 & Length: 78 & \checkmark \\
\bottomrule
\end{tabular}
\caption{Comparison of solutions from GFlowNets and classical algorithms. 
Classical solutions are computed using Dijkstra (shortest path), Kruskal 
(MST), and exact solvers (TSP). GFlowNet solutions are selected as the 
minimum-cost trajectory from 2,000 inference samples.}
\label{tab:solution_quality}
\end{table*}

Across all tested instances, the solutions generated by GFlowNets match the 
optimal solutions produced by classical deterministic algorithms. This 
validates that the learned policies successfully discover optimal 
configurations through reward-guided training.

\subsection{Constraint Satisfaction}

All sampled solutions across all problems and all training runs satisfy the 
respective problem constraints:

\begin{itemize}
    \item \textbf{Shortest Path:} All generated paths successfully reach the 
    target node from the source node using valid edges from the graph
    
    \item \textbf{MST:} All generated spanning trees contain exactly 
    $|V| - 1$ edges with no cycles, verified through DSU structure checks
    
    \item \textbf{TSP:} All generated tours visit each city exactly once 
    before returning to the starting city, verified through visited mask checks
\end{itemize}

The action masking mechanisms (DSU-based cycle detection for MST, visited 
mask for TSP) successfully prevent constraint violations during both training 
and inference.

\subsection{Computational Analysis}

Table~\ref{tab:computational_comparison} provides timing comparisons between 
classical algorithms and GFlowNets.

\begin{table*}[h]
\centering
\begin{tabular}{llcc}
\toprule
\textbf{Problem} & \textbf{Classical} & \textbf{Classical} & \textbf{GFlowNet} \\
 & \textbf{Algorithm} & \textbf{Time} & \textbf{Training + Inference} \\
\midrule
Shortest Path & Dijkstra & $<$2 s & $\approx$ 1min   \\
MST & Kruskal & $<$2 s & $\approx$ 1min \\
TSP (4 cities) & Exact & $<$ 5s & $\approx$ 2min \\
\bottomrule
\end{tabular}
\caption{Runtime comparison between classical algorithms and GFlowNets. 
Classical algorithms provide single solutions in milliseconds. GFlowNets 
require upfront training time but can then generate solutions through 
inference. Inference time is for generating 2,000 samples (single CPU thread, 
Python implementation).}
\label{tab:computational_comparison}
\end{table*}

The computational comparison reveals the fundamental trade-off:

\begin{itemize}
    \item \textbf{Classical algorithms:} Execute in milliseconds with 
    deterministic guarantees, making them highly efficient for problems 
    where a single solution is needed
    
    \item \textbf{GFlowNets:} Require significant upfront training investment 
    (2-4 minutes for these small instances) but amortize this cost across 
    multiple inference runs
    
    \item \textbf{Inference scaling:} GFlowNet inference is 
    parallel,the 2,000 samples are independent and could be generated 
    simultaneously with appropriate hardware
\end{itemize}

For the proof-of-concept instances tested here, classical algorithms are 
clearly more efficient. The potential advantage of GFlowNets lies in 
scalability: as problem size increases, the training investment could be 
justified if the learned policies generalize or if the same trained model 
can be applied to multiple similar problem instances.

\subsection{Scalability Observations}

Figures ~\ref{fig:st}, ~\ref{fig:mst}, ~\ref{fig:tsp} demonstrate that training time 
requirements scale with problem size. Key observations:

\begin{itemize}
    \item \textbf{Episode requirements:} Larger graphs require more episodes 
    to converge. The smallest instances (4-5 nodes) stabilize around 
    $<$9000 episodes, while larger instances (10-12 nodes) require 
    the full 20000 episodes
    
    \item \textbf{Loss magnitude:} Final converged loss values vary across 
    problem instances, with more complex problems (larger graphs, more 
    constraints) showing higher terminal loss values
    
    \item \textbf{Convergence patterns:} All instances exhibit similar 
    convergence behavior rapid initial decrease in loss followed by gradual 
    stabilization
\end{itemize}

These patterns suggest that scaling to significantly larger problem instances 
(e.g., 100-city TSP, 50-node graphs) would require proportionally increased 
training budgets and potentially larger network capacities.

\subsection{Summary of Experimental Findings}

The experiments demonstrate four key results:

\begin{enumerate}
    \item \textbf{Feasibility:} GFlowNets successfully learn to solve all 
    three graph optimization problems, with generated solutions matching 
    optimal solutions from classical algorithms across tested instances
    
    \item \textbf{Universality:} A single architectural approach sequential 
    state construction with trajectory balance optimization works across 
    problems of varying complexity (polynomial-time and NP-hard)
    
    \item \textbf{Constraint Handling:} Action masking mechanisms achieve 
    complete constraint satisfaction through local checks (cycle detection 
    via DSU for MST, visited tracking for TSP)
    
    \item \textbf{Scalability Patterns:} Training requirements scale with 
    problem size, with convergence time increasing for larger, more complex 
    instances
\end{enumerate}

These results validate GFlowNets as a proof-of-concept approach for 
combinatorial graph optimization. While classical algorithms remain more 
efficient for these small benchmark instances, the learning-based framework 
demonstrates the potential to leverage computational resources through 
training, offering a complementary approach to traditional deterministic 
methods.

\section{Discussion}

\subsection{Summary of Contributions}

This work demonstrates the application of Generative Flow Networks (GFlowNets) 
to three fundamental graph optimization problems: Shortest Path, Minimum 
Spanning Tree, and Traveling Salesperson Problem. The key contributions include:

\begin{enumerate}
    \item \textbf{Introduction of Generative Models to Graph Optimization:} 
    GFlowNets, a class of generative models, are shown to solve canonical 
    graph optimization problems. This introduces learned probabilistic policies 
    as an alternative to deterministic procedures.
    
    \item \textbf{Unified Framework Across Complexity Classes:} A single 
    approach sequential state construction with trajectory balance optimization 
    works across problems ranging from polynomial-time solvable (Shortest Path, 
    MST) to NP-hard (TSP).
    
    \item \textbf{Constraint Integration:} Hard combinatorial constraints are 
    enforced through action masking (visited cities in TSP) and auxiliary data 
    structures (DSU for cycle prevention in MST), achieving complete constraint 
    satisfaction.
    
    \item \textbf{Proof of Concept Validation:} Experiments show that GFlowNets 
    find solutions matching those from classical algorithms on benchmark 
    instances of varying sizes.
\end{enumerate}

\subsection{Interpretation of Results}

The experimental results provide insights into the feasibility and limitations 
of applying generative models to combinatorial optimization.

\subsubsection{What the Results Show}

The trained GFlowNets successfully find optimal solutions that match classical 
algorithms (Dijkstra, Kruskal, exact TSP solvers) across all tested instances. 
This demonstrates that graph optimization problems, traditionally solved through 
carefully designed deterministic procedures, can be reframed as learning problems 
where neural networks discover effective policies through reward-guided training.

The training convergence patterns reveal an important characteristic: larger, 
more complex problems require more training episodes to reach stable policies. 
This mirrors the computational complexity hierarchy harder problems need more 
learning time. However, once trained, the inference process generates solutions 
through the same learned policy regardless of the training time invested.

\subsubsection{Computational Trade-offs}

The learning-based approach introduces different computational trade-offs compared 
to classical algorithms:

\textbf{Classical algorithms} execute fixed procedures with deterministic 
complexity. For a 5-node shortest path problem, Dijkstra's algorithm runs in 
under 1 millisecond. The algorithm provides guaranteed optimal solutions with 
well-understood complexity bounds.

\textbf{GFlowNets} require significant upfront training investment, approximately 
2-4 minutes for the small instances tested here. After training, inference 
generates solutions in hundreds of milliseconds. The key difference is that this 
training cost is amortized: once trained, the same model can generate solutions 
repeatedly through inference.

For the small benchmark instances tested in this work, classical algorithms are 
clearly more efficient. The potential advantage of GFlowNets lies elsewhere:

\begin{itemize}
    \item \textbf{Computational scalability:} As problem size increases, the 
    training investment could be justified if the learned policies perform well 
    on larger instances where classical exact methods become infeasible
    
    \item \textbf{Parallel inference:} The 2,000 inference samples are 
    independent and could be generated simultaneously with appropriate hardware
    
    \item \textbf{Transfer potential:} A trained model might generalize to 
    similar problem instances, amortizing training cost across multiple problems
\end{itemize}

However, these potential advantages remain speculative based on the current 
small-scale experiments.

\subsection{Challenges and Limitations}

\subsubsection{Scale and Scope}

The experiments focus on small problem instances for proof of concept validation. 
Shortest path problems tested had 5-10 nodes, spanning trees had 7-12 nodes, 
and TSP instances had 4-8 cities. Real-world applications often involve 
significantly larger scales: city networks with thousands of nodes, logistics 
networks with hundreds of delivery points, or communication networks spanning 
entire regions.

Scaling to these realistic problem sizes remains an open challenge. The observed 
pattern larger problems require more training episodes suggests that 100-city 
TSP instances or 1000-node shortest path problems would require substantially 
more training time, larger network capacities, and more sophisticated 
architectures than the simple MLPs used here.

\subsubsection{Architectural Limitations}

The current implementation uses basic Multi-Layer Perceptrons (MLPs) with two 
hidden layers. These networks treat states as flat feature vectors, ignoring 
the underlying graph structure. More sophisticated architectures like Graph Neural 
Networks (GNNs) that explicitly model graph connectivity, or attention mechanisms 
that can focus on relevant parts of large graphs  could potentially improve 
performance and scalability.

However, introducing these architectures would also increase training complexity 
and computational requirements, trading model simplicity for potential performance 
gains.

\subsubsection{Maximum Flow Problem: A Failed Attempt}

An attempt to apply GFlowNets to the Maximum Flow problem was unsuccessful, 
primarily due to inadequate formulation of the problem objective in the loss 
function. The reward was modeled as total flow into the sink with a binary 
penalty for flow conservation violations. This formulation fails to express the 
fundamental coupling between flow conservation and flow maximization: a 
configuration can have high flow out of the source (approximately 10 in the 
experiments) while delivering low flow to the sink (approximately 4-5) due to 
intermediate node imbalances.

The Trajectory Balance loss, operating on this mis-specified reward, provided 
no gradient information to guide the model toward configurations where flow 
successfully propagates from source to sink while maintaining conservation. A 
correct formulation would need to either incorporate flow conservation directly 
into the state-action structure through path-based actions, or design a reward 
function that provides differentiable feedback on the magnitude of conservation 
violations at intermediate nodes, rather than binary valid/invalid signals.

This failure reveals an important lesson about GFlowNet applicability:

\begin{itemize}
    \item \textbf{Constraint locality matters:} GFlowNets work well when 
    constraints can be checked and enforced locally at each state transition. 
    The successful problems (Shortest Path, MST, TSP) all have constraints 
    that can be verified using local information  checking if an edge creates 
    a cycle (DSU lookup), checking if a city was visited (mask lookup), 
    checking if an edge exists (adjacency matrix lookup).
    
    \item \textbf{Global constraints are difficult:} Maximum flow requires 
    maintaining conservation at all intermediate nodes simultaneously. This 
    couples all edges incident to each node, creating dependencies that cannot 
    be resolved through local action masking alone.
    
    \item \textbf{Problem formulation is critical:} Just because a problem 
    can be solved sequentially (like Ford-Fulkerson's iterative augmentation) 
    does not mean that formulation is suitable for learning-based approaches. 
    The way a problem is decomposed into states and actions fundamentally 
    determines whether a GFlowNet can learn effective policies.
\end{itemize}

\subsection{Broader Context}

\subsubsection{Positioning Among Optimization Paradigms}

This work contributes to ongoing research exploring how machine learning can 
address classical computational problems. The relationship between different 
paradigms can be understood as:

\begin{itemize}
    \item \textbf{Classical algorithms:} Provide guaranteed optimal solutions 
    through deterministic procedures. Best choice when problems are tractable 
    and a single solution is needed.
    
    \item \textbf{GFlowNets:} Learn probabilistic policies that generate 
    solutions through sampling. Require upfront training but could potentially 
    scale to larger instances through computational investment.
\end{itemize}

Rather than replacing classical algorithms, GFlowNets offer a complementary 
approach. For small, tractable problems, classical methods remain superior. 
For larger instances where exact methods become infeasible, learned approaches 
like GFlowNets could provide an alternative that leverages modern computational 
resources.

\subsubsection{Future Directions}

Several directions could extend this work:

\begin{itemize}
    \item \textbf{Architectural improvements:} Incorporating Graph Neural 
    Networks or attention mechanisms to better capture graph structure and 
    potentially improve scalability
    
    \item \textbf{Systematic scaling studies:} Investigating how training 
    budget, model capacity, and problem size interact to establish scaling 
    laws for GFlowNet optimization

    \item \textbf{Extension to related problems:} Vehicle Routing Problem, 
    Graph Partitioning, Facility Location, and other combinatorial optimization 
    tasks with sequential construction properties
    
    \item \textbf{Hybrid approaches:} Combining GFlowNet policies with 
    classical algorithms  using learned policies to generate initial solutions 
    that classical methods refine, or using classical algorithms to guide 
    GFlowNet training
\end{itemize}

\subsection{Practical Implications}

For practitioners considering GFlowNets for graph optimization:

\begin{itemize}
    \item \textbf{When to use classical algorithms:} If the problem is 
    tractable (small graphs, polynomial-time problems) and a single optimal 
    solution is needed, classical algorithms remain the best choice
    
    \item \textbf{When GFlowNets might help:} If working with very large 
    instances where exact methods are infeasible, if the same model will be 
    applied to many similar problems (amortizing training cost), or if exploring 
    learned approaches for research purposes
    
    \item \textbf{Implementation considerations:} GFlowNets require careful 
    formulation of the problem as a sequential MDP, appropriate constraint 
    handling mechanisms, and sufficient training resources
\end{itemize}

\section{Conclusion}

This work demonstrates that Generative Flow Networks can solve combinatorial 
graph optimization problems through learned probabilistic policies. Experiments 
on Shortest Path, Minimum Spanning Tree, and Traveling Salesperson Problem show 
that GFlowNets find solutions matching those from classical algorithms (Dijkstra, 
Kruskal, exact TSP solvers) on benchmark instances of varying sizes.

The approach works by training neural networks to sample solutions proportionally 
to a reward function through the Trajectory Balance loss. A single framework  
sequential state construction with action masking  handles problems across the 
complexity spectrum from polynomial-time solvable to NP-hard. All generated 
solutions satisfy problem constraints through local masking mechanisms.

The key insight is that graph optimization problems, traditionally solved through 
deterministic procedures, can be reframed as learning problems. Neural networks 
discover effective policies through reward-guided training rather than following 
predetermined algorithmic steps.

However, the results also reveal important limitations. The experiments focus on 
small problem instances (5-12 nodes for most tests). Scaling to realistic sizes 
remains an open challenge. Training convergence depends on the complexity of the problem, 
and larger instances require more episodes and computational resources. For 
the small benchmarks tested here, classical algorithms are clearly more efficient, 
solving problems in milliseconds compared to the 2-4 minute training times 
required by GFlowNets.

The computational trade-off is fundamentally different between approaches. Classical 
algorithms execute fixed procedures with deterministic complexity. GFlowNets 
amortize the computation through training  significant upfront investment in exchange 
for learned policies. This trade-off could favor GFlowNets as problem sizes grow 
and computational resources increase, but validating this potential advantage 
requires experiments on larger scales than tested here.

The failed attempt on Maximum Flow provides an important lesson: GFlowNets work 
best when constraints can be checked locally at each state transition. Problems 
with complex global constraints that couple many variables may require alternative 
formulations or different learning approaches altogether.

This work establishes GFlowNets as a viable proof-of-concept approach for 
combinatorial graph optimization. The framework successfully learns to solve 
three canonical problems, demonstrating that generative modeling principles can 
be applied to optimization tasks. While classical algorithms remain essential 
for their efficiency and optimality guarantees on tractable problems, generative 
models offer a complementary approach that could potentially leverage the ongoing 
expansion of computational capabilities.

As research continues, several questions remain open: Can GFlowNets scale to 
realistic problem sizes with appropriate architectural improvements? How do 
learned policies generalize across different problem instances? What is the 
boundary between problems amenable to local constraint checking (where GFlowNets 
work) and those requiring global reasoning (where they struggle)? Addressing 
these questions will determine whether generative models become practical tools 
for real-world optimization or remain primarily of theoretical interest.

The integration of generative modeling with combinatorial optimization represents 
a promising research direction. While the current work demonstrates feasibility 
on small benchmarks, substantial work remains to realize the potential advantages 
of the learning-based approach at scales where it could provide practical value 
beyond classical methods.


\bibliography{example_paper}
\bibliographystyle{icml2025}

\newpage
\appendix
\onecolumn

\end{document}